\documentclass[runningheads]{llncs}
\usepackage[T1]{fontenc}
\usepackage{framed,multirow}
\usepackage{amssymb,bm}
\usepackage{latexsym}
\usepackage{booktabs}
\usepackage{xcolor}
\usepackage{amsmath}
\usepackage{algorithm}
\usepackage{algorithmic}
\usepackage{enumitem}
\usepackage{graphicx,verbatim}
\begin{document}
\title{Recognizing Surgical Phases Anywhere: Few-Shot Test-time Adaptation and Task-graph Guided Refinement}

\titlerunning{SPA: Surgical Phase Anywhere}
\author{Kun Yuan\inst{1,3,7} \and 
Tingxuan Chen\inst{3} \and 
Shi Li\inst{1} \and 
Joel L. Lavanchy\inst{4} \and 
Christian Heiliger\inst{5} \and 
Ege \"{O}zsoy\inst{3} \and 
Yiming Huang\inst{6} \and 
Long Bai\inst{6} \and 
Nassir Navab\inst{3} \and 
Vinkle Srivastav \inst{1,2} \and 
Hongliang Ren\inst{6} \and 
Nicolas Padoy\inst{1,2} 
}

\authorrunning{Kun. et al.}
%

\institute{University of Strasbourg, CNRS, INSERM, ICube, UMR7357, Strasbourg, France \and IHU Strasbourg, Strasbourg, France \and
CAMP, Technische Universit\"at M\"unchen, Munich, Germany \and University Digestive Health Care Center – Clarunis, 4002 Basel, Switzerland \and Ludwig Maximilian University of Munich, Munich, Germany \and Chinese University of Hong Kong, Hong Kong SAR, China \and Munich Center for Machine Learning, Germany \email{kyuan@unistra.fr;npadoy@unistra.fr}}

\maketitle              
\begin{abstract}
The complexity and diversity of surgical workflows, driven by heterogeneous operating room settings, institutional protocols, and anatomical variability, present a significant challenge in developing generalizable models for cross-institutional and cross-procedural surgical understanding. While recent surgical foundation models pretrained on large-scale vision-language data offer promising transferability, their zero-shot performance remains constrained by domain shifts, limiting their utility in unseen surgical environments. To address this, we introduce \textbf{S}urgical \textbf{P}hase \textbf{A}nywhere (SPA), a lightweight framework for versatile surgical workflow understanding that adapts foundation models to institutional settings with minimal annotation. SPA leverages few-shot spatial adaptation to align multi-modal embeddings with institution-specific surgical scenes and phases. It also ensures temporal consistency through diffusion modeling, which encodes task-graph priors derived from institutional procedure protocols. Finally, SPA employs dynamic test-time adaptation, exploiting the mutual agreement between multi-modal phase prediction streams to adapt the model to a given test video in a self-supervised manner, enhancing the reliability under test-time distribution shifts. SPA is a lightweight adaptation framework, allowing hospitals to rapidly customize phase recognition models by defining phases in natural language text, annotating a few images with the phase labels, and providing a task graph defining phase transitions. The experimental results show that the SPA framework achieves state-of-the-art performance in few-shot surgical phase recognition across multiple institutions and procedures, even outperforming full-shot models with 32-shot labeled data.


\end{abstract}
\footnote{\small \textit{This manuscript has been accepted for publication and will be included in the proceedings of MICCAI 2025.}}

\section{Introduction}
\label{sec:introduction}

\begin{figure*}
  \centering
  \includegraphics[width=\columnwidth]{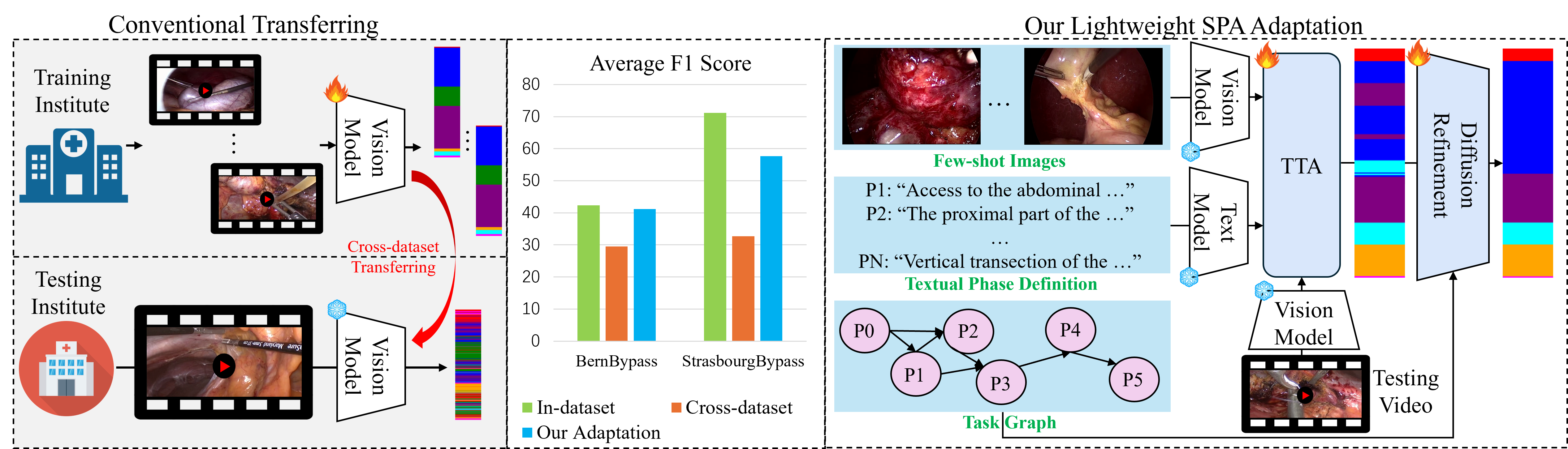} 
  \caption{Conventional adaptation struggles with performance drops across institutions due to spatial and procedural variations. Our institution-specific adaptation minimizes manual effort by requiring only textual phase definitions, a few labeled images, and a task graph from the target hospital, to adapt spatial understanding and enforce institution-specific temporal constraints.}
  \label{fig1}
\end{figure*}

Deep learning models have made significant progress in the field of surgical data science~\cite{maier2017surgical,yuan2025learning}, particularly in surgical phase recognition~\cite{czempiel2020tecno,wang2022autolaparo,garrow2021machine}, which is essential for improving patient safety~\cite{mascagni2022computer}, reducing errors, and optimizing communication in the operating room (OR)~\cite{ozsoy2024oracle}. Surgical phase recognition can provide real-time insights into procedure progress, enabling early anomaly detection~\cite{murali2022latent} and context-aware decision support. However, current phase recognition models face limitations when deployed in real clinical settings. Specifically, their performance often degrades in unseen institutions and procedures due to variability in surgical settings, such as different procedure protocols, OR setups~\cite{ozsoy2024oracle}, instrumentation~\cite{lavanchy2024challenges,jaspers2025scaling,wang2024video}, and variations in patient anatomies~\cite{honarmand2024vidlpro}, as shown in Fig.~\ref{fig1}. This variability necessitates costly re-annotation and retraining for each new clinical site, which hinders the scalability of phase recognition models. Therefore, there is an urgent need for generalizable and lightweight adaptation methods that can bridge these gaps without requiring extensive re-annotation or retraining, ensuring broader real-world applicability. 

Few-shot transfer learning~\cite{shakeri2024few,silva2024closer,zhang2021tip,zhou2025ultraad} with foundation models~\cite{radford2021learning,yuan2024hecvl} is a promising approach to address domain shifts in surgical tasks, particularly variations in instruments and anatomies. By leveraging pretrained representations from large-scale datasets such as SVL-Pretrain~\cite{yuan2025learning} and fine-tuning with a few labeled samples, it reduces reliance on extensive annotations. However, existing few-shot studies rely on fine-tuning with a significant fraction of data (e.g., 12.5\% or 25\% in~\cite{ramesh2023dissecting}), requiring hundreds to thousands of labeled samples. This assumption is impractical in clinical applications. Standard few-shot learning~\cite{silva2024closer,zhang2021tip} with K-shot N-class settings often overfits on surgical video datasets, which are limited compared to radiology~\cite{wang2022medclip,wu2023medklip,ma2025eye}, histology~\cite{huang2023visual,ikezogwo2024quilt}, or ophthalmology~\cite{silva2025foundation,hu2024ophclip}. Each surgical video is typically from one patient, resulting in sparse and unrepresentative training samples. Consequently, hyperparameter tuning on small validation sets from a single surgical video fails to capture dataset diversity, resulting in poor generalization under test-time distribution shifts.

Current few-shot learning methods~\cite{shakeri2024few,silva2024closer,zhang2021tip} focus on spatial, image-based adaptation and fail to capture temporal dependencies essential for surgical workflow modeling, which vary across institutions. For instance, different hospitals follow distinct procedural protocols for gastric bypass surgery~\cite{lavanchy2024challenges}. Sparse, isolated frames lack the continuity needed to model such variations, making existing methods ineffective for cross-institutional generalization. To address this, we leverage institutional-specific task graphs as temporal priors to learn phase transition distributions using a generative approach, i.e., diffusion modeling.

We introduce Surgical Phase Anywhere (SPA), a lightweight adaptation framework enabling foundation models to generalize across institutions with minimal supervision by integrating spatiotemporal priors. SPA enhances spatial generalization via few-shot adaptation, training a lightweight linear layer on a frozen foundation model to align multi-modal embeddings using limited labeled images and textual phase descriptions. To address temporal variability, SPA employs diffusion adaptation module that encodes institutional task-graph priors, ensuring phase predictions align with the target institution’s protocol. Additionally, SPA uses self-supervised test-time adaptation, leveraging multi-modal mutual agreement to refine predictions during inference, adapting to distribution shifts from procedural variations. SPA enables scalable, data-efficient surgical workflow modeling through spatial, temporal, and test-time adaptation.

Our key contributions are: (1) \textbf{S}urgical \textbf{P}hase \textbf{A}nywhere (SPA), an institution-specific adaptation framework that transfers surgical vision-language foundation models to phase recognition across procedures and institutions with minimal supervision, (2) a spatiotemporal adaptation strategy, integrating spatial priors via few-shot learning and temporal priors via diffusion modeling, (3) test-time adaptation, enhancing robustness against test-time distribution shifts through multi-modal mutual agreement, and (4) extensive evaluations, showing strong generalization, sometimes surpassing full-shot methods with $200\times$ less data.

\section{Method}
\label{sec:method}

\begin{figure*}[!htb]
  \centering
  \includegraphics[width=0.98\columnwidth]{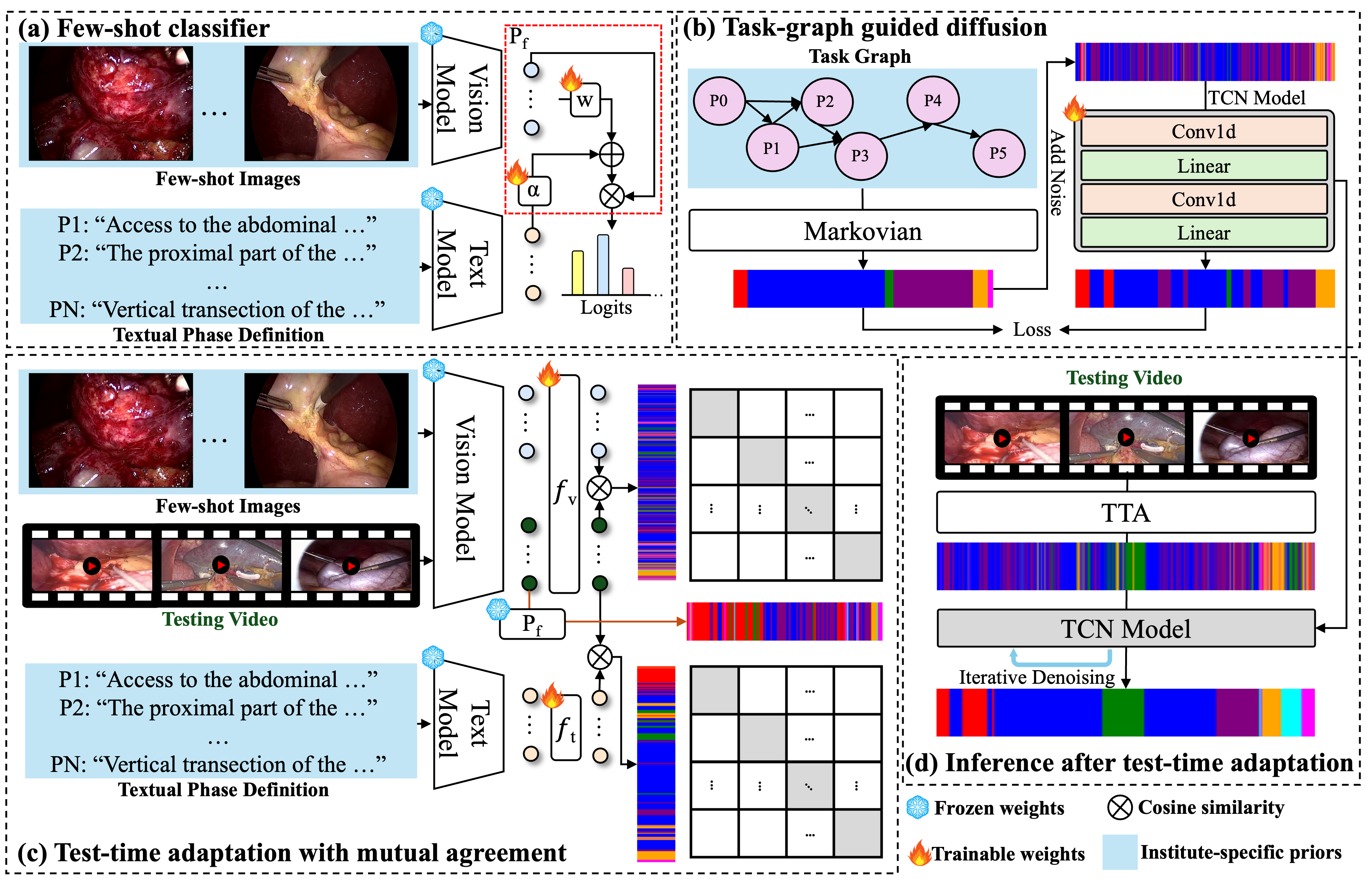} 
  \caption{Pipeline of SPA. During training, we first train (a) few-shot classifier for spatial adaptation, and (b) task-graph guided diffusion model for temporal adaptation. During inference, TTA updates model parameters using self-supervision, followed by the diffusion model to ensure temporally coherent phase transitions.}
  \label{fig2}
\end{figure*}

SPA has three modules: spatial adaptation via few-shot learning (Sec.\ref{sec:few_shot}), temporal adaptation via task-graph guided diffusion modeling (Sec.\ref{sec:diffusion}), and test-time adaptation (Sec.~\ref{sec:TTA}). We follow the standard few-shot learning definition, where N-shot means using N × K labeled samples, with K samples per class.

\subsection{Spatial adaptation via few-shot learning}
\label{sec:few_shot}

As shown in Fig.~\ref{fig2} (a), we first build a few-shot classifier using the target institution-specific priors, i.e., textual phase class descriptions from each center and a few-shot set of labeled images. For each image $\bm{x}_i$, its vision embedding is computed as: $\bm{f}_{i} = \bm{\theta}_{v}(\bm{x}_{i})$ using the the frozen pre-trained visual encoder $\theta_{v}$. For each phase class $k \in \{1, \dots, K\}$, we encode its textual description $\bm{d}_{k}$ (e.g., ``Preparation: we introduce trocar and ...'') using a frozen pretrained text encoder: $\bm{t}_k = \bm{\theta}_{t}(\bm{d}_{k}).$ Following the work of Shakeri et al.~\cite{shakeri2024few}, we then learn a text-driven linear classifier ${P_f}$ consisting of vision class prototypes $\mathbf{w} = (\bm{w}_k)_{1\leq k \leq K}$ blended with text embeddings via learnable multipliers $\boldsymbol{\alpha} = (\alpha_k)_{1\leq k \leq K} $:
\begin{equation}
    \label{CE-loss}
    p_{ik}(\mathbf{w}, \boldsymbol{\alpha}) = \frac{\exp \left ( \bm{f}_{i}^{t} (\bm{w}_{k} + \alpha_k \bm{t}_{k}) \right) }{\sum_{j=1}^{K}\exp \left (\bm{f}_{i}^{t} (\bm{w}_{j} + \alpha_j \bm{t}_{j}) \right )}.
\end{equation}
During training, only the weights of the class prototypes in linear layer $\mathbf{w}$ and the learnable multiplier $\boldsymbol{\alpha}$ are optimized via gradient descent, while text embeddings $\bm{t}_k$ remain fixed. The learning objective is cross-entropy loss to learn to classify each few-shot image into one of the phase classes.

\subsection{Temporal adaptation via task-graph guided diffusion}
\label{sec:diffusion}

The initial non-temporal predictions can be noisy and inaccurate. To enforce temporal coherence and improve prediction accuracy, we introduce a task-graph guided diffusion model. The institution-specific task graph generates synthetic phase sequences, which are then used to train the diffusion model, enabling it to capture the contextual relationships between different phases. During inference, the trained diffusion model refines the noisy coarse phase predictions to align with the specific procedural protocols of the institution.

\textbf{Synthetic data generation from task graph:} First, we synthesize temporally coherent surgical phase sequences $X \in \mathbb{R}^{L}$ by encoding phase transition graph, i.e., Task Graph, $\mathcal{G} = (V,E)$. The task graph represents the valid transitions between surgical phases. Here, $V$ is the set of graph nodes $v_j$ to represent surgical phases, and $E$ is the set of directed edges $e_{ij}$ indicating permissible phase transitions. The edges encode institution-specific phase dependencies, ensuring that generated sequences follow target procedure protocols. For each surgical phase node, we also have an estimated time of their temporal bounds ($[L_{\min}^i, L_{\max}^i]$). Therefore, we can synthesize surgical phase sequence using a Markovian process using the following:
\begin{equation}
    P(X_{l+1}|X_l) =
\begin{cases}
    \mathcal{U}({v_j \in V \mid e_{ij} \in E}) & \text{if } \Delta l^i < L_{\max}^i \\
    0 & \text{otherwise}
\end{cases}.
\label{markovian}
\end{equation}

This equation dictates that the next phase $X_{l+1}$ is uniformly sampled from the set of valid successor phases in the task graph ($\mathcal{U}({v_j \in V \mid e_{ij} \in E})$), and the current phase duration has not exceeded its upper bound ($L_{\max}^i$). The synthetic sequences generated using Eq.~\ref{markovian} are representative of the target institution’s surgical phase transition distribution. Also, the Markovian generation process enables the creation of diverse yet valid sequences, effectively modeling variations in real-world procedures.

\textbf{Diffusion model in hidden state space:} We train a diffusion model on the generated synthetic phase sequences $X$ to learn the phase transition distribution specific to the target institution as shown in Fig.~\ref{fig2}(b). The model operates in a \textit{hidden state space} \(H \subseteq \mathbb{R}^{C}\), where each hidden state \({h}_l \in H\) represents the phase label at timestamp \(l\). These hidden states are mapped to phase probabilities: $ H = \left\{ {h}_1, \dots, {h}_L \right\}  \sigma({h}_l) \in \mathbb{R}^C,$ where $\sigma({h}_l)$ is the one-hot feature vector.

\textbf{Forward process in hidden space.} The forward process gradually introduces noise into the phase sequences, disrupting their structure. This helps the model learn to recover meaningful phase transitions from noisy data. Formally, the forward process is defined as:
\begin{equation}
q({H}_t | {H}_{t-1}) = \mathcal{N}\left({H}_t; \sqrt{1 - \beta_t} {H}_{t-1}, \beta_t {I} \right) ,
\end{equation} $\beta_t$ is the noise variance at step $t$. Here, $\mathcal{N}(\mu, \Sigma)$ represents a multivariate normal (Gaussian) distribution with mean $\mu$ and covariance $\Sigma$. This means that at each step $t$, the hidden state $H_t$ is sampled from a Gaussian distribution.

\textbf{Reverse process with surgical priors.} The reverse process learns to denoise sequences to learn the temporal priors from the synthetic data:
\begin{equation}
    p_\theta({H}_{t-1} | {H}_t) = \mathcal{N}\left({H}_{t-1}; \boldsymbol{\mu}_\theta({H}_t, t), \boldsymbol{\Sigma}_\theta({H}_t, t) \right) ,
\end{equation}
where $\boldsymbol{\mu}_\theta$ and $\boldsymbol{\Sigma}_\theta$ are learned parameters. During training, noise is progressively added to phase sequences, and the reverse process learns to denoise them, refining noisy predictions into surgically plausible sequences. During inference, a coarse phase prediction and noise step are input, and the reverse process refines it iteratively to align with learned temporal priors, i.e., procedure protocols.

\subsection{Test-time adaptation (TTA) with mutual agreement}
\label{sec:TTA}

Surgical foundation models generate multi-modal embeddings, enabling image-text comparisons via cosine similarity. In the SPA framework, each test image produces three prediction streams: (1) \textbf{Reference prediction}, which compares the test image to few-shot reference images (Sec.~\ref{sec:few_shot}), given by \( S_{ref} = \sigma(V \cdot R^\top) C \), where \( V \) is the test image embedding, \( R \) the reference visual embeddings, and \( C \) is their class associations; (2) \textbf{Vision-language prediction}, which matches the image to phase descriptions using \( S_{vl} = \sigma(V \cdot T^\top) \); and (3) \textbf{Few-shot prediction}, obtained from a previously trained few-shot classifier (Sec.~\ref{sec:few_shot}), given by \( S_{fs} \).

These predictions may individually generalize poorly due to test-time distribution shifts. However, ensembling them provides complementary information that if they co-agree on a prediction, it is likely correct. To leverage this mutual agreement, we introduce a self-supervised learning strategy for test-time adaptation. We add two linear layers $f_v$ and $f_t$ on top of the vision and text encoders to refine embeddings, thus $
{S}_{vl} = \sigma(f_v(V) \cdot f_t(T)^\top/\tau)$ and ${S}_{ref} = \sigma(f_v(V) \cdot f_v(R)^\top){C}.$

To align predictions across modalities, we define a contrastive loss:  
\[
\mathcal{L} = \mathcal{L}_{mutual}(S_{ref}, S_{vl}) + \mathcal{L}_{mutual}(S_{fs}, S_{ref}),
\] where $
\mathcal{L}_{mutual}(A, B) = -\frac{1}{L} \sum_{l=1}^L \sum_{k=1}^K A_{l,k} \log B_{l,k}$
penalizes frame-wise mismatches by aligning the diagonal elements of similarity matrices, enforcing cross-modal consistency at each timestamp. Overall, our SPA inference has two steps: first, we adapt the vision and text linear layers using \(\mathcal{L}\) to refine predictions for the test video. Then, we fuse the three prediction streams and further refine them using the previously trained diffusion model (Sec.~\ref{sec:diffusion}).

\section{Experiment setup}
\label{sec:experiments_setup}

\textbf{Target procedures and institutions:} Evaluation includes diverse surgical phase recognition datasets spanning various surgical procedures. \textbf{Cholecystectomy:} We utilize Cholec80~\cite{twinanda2016endonet} as it represents a well-established and routinely performed surgical procedure. This dataset requires anatomical understanding, including structures such as the gallbladder and liver; \textbf{Gastric Bypass:} To evaluate cross-dataset generalizability, we consider both StrasbourgBypass~\cite{lavanchy2024challenges} and BernBypass~\cite{lavanchy2024challenges}. Despite both datasets featuring gastric bypass surgeries, differences in phase definitions and procedural protocols create a significant domain gap, requiring adaptation of surgical foundation models. \textbf{Hysterectomy}: We also evaluate laparoscopic hysterectomy phase recognition using Autolaparo~\cite{wang2022autolaparo}.

\textbf{Implementation details:} This work adapts a generalist foundation model for institution-specific surgical phase recognition. We use PeskaVLP~\cite{yuan2024procedure}, an open-access surgical vision-language foundation model pretrained on surgical video lectures~\cite{yuan2025learning}. To retain pretraining knowledge and minimize computational costs, PeskaVLP’s encoders remain frozen during our adaptation. Few-shot adaptation (Sec.\ref{sec:few_shot}) is conducted on an RTX 3090 with a learning rate of 0.01. The task-graph-guided diffusion model (Sec.\ref{sec:diffusion}) is trained on a V100 GPU for 20 epochs with a batch size of 128 and a learning rate of 0.0001. Test-time adaptation (Sec.~\ref{sec:TTA}) runs on an RTX 3090 for 15 epochs with a 0.0001 learning rate, processing a 30-minute video in 22 seconds. Inference estimates noise levels and applies the diffusion model, requiring just 0.26 seconds for a 30-minute video.

\section{Results and discussions}

\begin{table}[t!]
  \scriptsize
    \caption{Comparison of few-shot methods using average F1 (\%) across four benchmarks. For each N-shot K-class setting, results are averaged over three subsets. The RN50 row is populated with values from SOTA papers, Cholec80~\cite{alapatt2024jumpstarting}, Bypass~\cite{lavanchy2024challenges}.} \label{table:main}
    \resizebox{1\textwidth}{!}{
    \begin{tabular}{cccccc}
        \toprule
        Methods & Shots & Cholec80 & StrasBypass & BernBypass & Autolaparo\\
        \midrule
        CLIP~\cite{radford2021learning} & 0 & 13.1 & 5.5 & 4.1 & 18.3 \\
        PeskaVLP~\cite{yuan2024procedure} & 0 & 34.2 & 28.6 & 22.6 & 28.6 \\
        RN50 & Full & 80.3 & 71.1 & 42.4 & NA \\
        \midrule

        Tip-Adapter-F~\cite{zhang2021tip}& 1 & $29.29_{\pm{}0.53}$ & $14.88_{\pm{}0.03}$ & $14.29_{\pm{}0.00}$ & $26.58_{\pm{}0.00}$\\
        Linear probe (LP)~\cite{shakeri2024few} & 1 & $26.19_{\pm{}6.75}$ & $24.70_{\pm{}3.64}$ & $14.95_{\pm{}1.35}$ & $33.90_{\pm{}5.19}$\\
        LP+text~\cite{shakeri2024few} & 1 & $32.21_{\pm{}1.88}$ & $23.53_{\pm{}3.59}$ & $14.12_{\pm{}2.36}$ & $32.80_{\pm{}5.81}$\\
        SPA & 1 & $44.53_{\pm{}5.77}$ & $30.68_{\pm{}5.24}$ & $24.31_{\pm{}4.29}$ & $51.48_{\pm{}6.26}$\\
        \bottomrule
        
        \toprule
        Tip-Adapter-F~\cite{zhang2021tip}& 16 & $40.75_{\pm{}2.15}$ & $34.29_{\pm{}2.12}$ & $24.54_{\pm{}1.97}$ & $37.51_{\pm{}1.31}$\\
        Linear probe (LP)~\cite{shakeri2024few} & 16 & $39.06_{\pm{}1.04}$ & $38.48_{\pm{}1.88}$ & $27.36_{\pm{}0.45}$ & $42.60_{\pm{}2.78}$\\
        LP+text~\cite{shakeri2024few} & 16& $38.24_{\pm{}2.33}$ & $35.42_{\pm{}1.35}$ & $25.01_{\pm{}0.50}$ & $39.43_{\pm{}4.83}$\\
        SPA & 16 & $55.05_{\pm{}4.80}$ & $52.39_{\pm{}1.88}$ & $38.73_{\pm{}1.58}$ & $58.91_{\pm{}4.72}$\\
        \bottomrule

        \toprule
        Tip-Adapter-F~\cite{zhang2021tip}& 32 & $42.05_{\pm{}1.04}$ & $35.83_{\pm{}4.08}$ & $26.98_{\pm{}1.68}$ & $39.75_{\pm{}2.11}$\\
        Linear probe (LP)~\cite{shakeri2024few} & 32 & $38.37_{\pm{}1.37}$ & $42.79_{\pm{}0.61}$ & $29.74_{\pm{}0.85}$ & $44.72_{\pm{}1.25}$\\
        LP+text~\cite{shakeri2024few} & 32 & $38.36_{\pm{}1.81}$ & $40.35_{\pm{}1.25}$ & $26.01_{\pm{}0.79}$ & $39.40_{\pm{}3.35}$\\
        SPA & 32 & $55.69_{\pm{}3.99}$ & $55.80_{\pm{}1.32}$ & $43.54_{\pm{}1.96}$ & $60.36_{\pm{}3.14}$\\
        \bottomrule
\end{tabular}
 }
\end{table}

\begin{figure*}[t!]
  \centering
  \includegraphics[width=1\columnwidth]{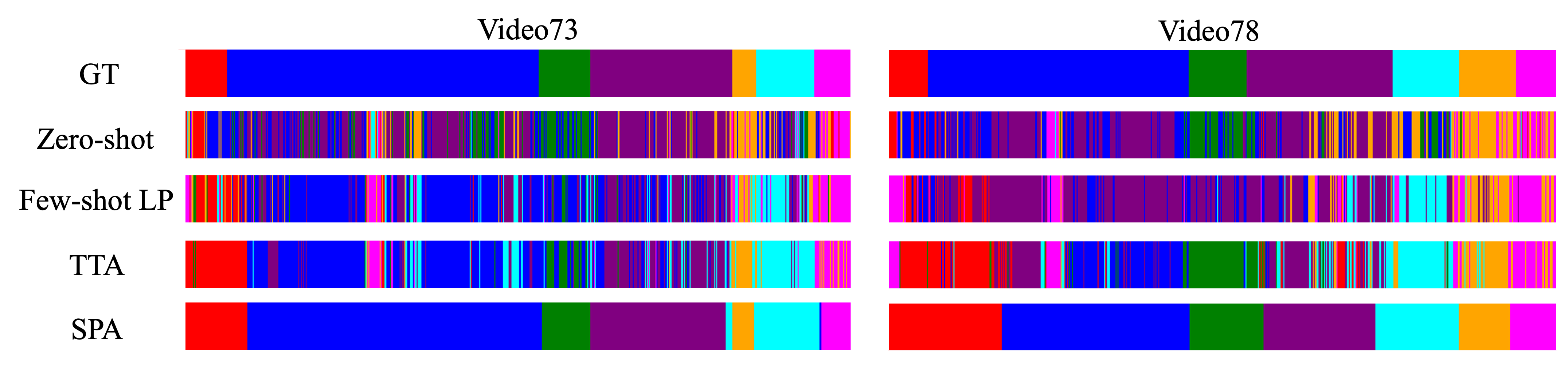} 
  \caption{32-shot phase recognition shows SPA reduces noise and fragmentation.}
  \label{fig3}
\end{figure*}

Tab.~\ref{table:main} shows that SPA effectively adapts across diverse surgical procedures (e.g., cholecystectomy vs. gastric bypass) and institutions (e.g., StrasBypass vs. BernBypass) by leveraging general knowledge from pretrained vision-language model~\cite{yuan2024procedure}. By leveraging institutional-specific spatiotemporal priors, SPA surpasses the SOTA few-shot methods, e.g., improving F1 by $+17.33\%$ on Cholec80 ($32$-shot). This scalability highlights its ability to refine domain-agnostic pretrained features with minimal institution-specific data. SPA even achieves a $43.54\%$ F1 on BernBypass ($32$-shot), surpassing the full-shot method ($42.4\%$) with $300\times$ less labeled data, demonstrating its viability for real-world settings where labeled data is scarce. As shown in Fig.~\ref{fig3}, SPA reduces fragmentation and errors in phase predictions. Both zero-shot and few-shot LP predictions exhibit high fragmentation, with disordered phase transitions and weak temporal consistency. TTA improves phase prediction accuracy and reduces fragmentation, but the predictions remain noisy. SPA further enhances recognition, yielding predictions closest to the ground truth with clear phase boundaries.

\begin{table*}[t!]
\centering
\caption{Ablation study. We conduct surgical phase recognition and report F1 Score. TTA: test-time adaptation. TG: Task-graph guided diffusion modeling.}
\label{table:ablation}
\begin{tabular}{l|c|c|c|c|c}
\toprule
& Shots & LP+text & +TTA & +TTA +TG-In & +TTA +TG-Cross \\
\midrule
\multirow{2}{*}{Cholec80} 
& 16 & $38.24 _{\pm{}\!2.33}$ & $51.89 _{\pm{}\!4.39}$ & $55.05 _{\pm{}\!4.80}$ & NA \\
& 32 & $38.36 _{\pm{}\!1.81}$ & $51.23 _{\pm{}\!2.93}$ & $55.69 _{\pm{}\!3.99}$ & NA \\
\midrule
\multirow{2}{*}{StrasBypass} 
& 16 & $35.42 _{\pm{}\!1.35}$ & $41.70 _{\pm{}\!1.60}$ & $52.39 _{\pm{}\!1.88}$ & $39.09 _{\pm{}\!3.01}$\\
& 32 & $40.35 _{\pm{}\!1.25}$ & $43.32 _{\pm{}\!0.38}$ & $55.80 _{\pm{}\!1.32}$ & $42.86 _{\pm{}\!1.35}$\\
\midrule
\multirow{2}{*}{BernBypass} 
& 16 & $25.01 _{\pm{}\!0.50}$ & $28.72 _{\pm{}\!3.66}$ & $38.73 _{\pm{}\!1.58}$ & $41.64 _{\pm{}\!1.05}$\\
& 32 & $26.01 _{\pm{}\!0.79}$ & $29.09 _{\pm{}\!0.62}$ & $43.54 _{\pm{}\!1.96}$ & $45.38 _{\pm{}\!1.78}$\\
\midrule
\multirow{2}{*}{Autolaparo} 
& 16 & $39.43_{\pm{}4.83}$ & $47.59_{\pm{}3.42}$ & $58.91_{\pm{}4.72}$ & NA \\
& 32 & $39.40_{\pm{}3.35}$ & $50.68_{\pm{}1.07}$ & $60.36_{\pm{}3.14}$ & NA \\
\bottomrule
\end{tabular}
\end{table*}

\subsection{Ablation study}  
We conduct an ablation study in Tab.~\ref{table:ablation} to evaluate the effect of our proposed modules, i.e., test-time adaptation (TTA), task-graph diffusion modeling (TG). \textbf{Test-time adaptation:} TTA significantly enhances model generalization for unseen test videos. Across all datasets, TTA improves F1 scores by notable margins, e.g., $+13.65\%$ on Cholec80 ($16$-shot) and $+6.28\%$ on StrasBypass ($16$-shot). These results highlight the effectiveness of self-supervised adaptation in mitigating distribution shifts, enabling more robust phase recognition in varied surgical environments. \textbf{Task-graph guided diffusion:} We assess the transferability of task graphs using institution-specific (TG-In) and cross-institutional (TG-Cross) priors. As shown in Tab.~\ref{table:ablation}, TG-In significantly boosts performance, improving StrasBypass by $+16.97\%$ (16-shot) and BernBypass by $+13.72\%$. However, TG-Cross shows mixed results: applying StrasBypass’s task graph enhances BernBypass, while BernBypass’s task graph reduces StrasBypass performance. This shows that procedural differences impact effectiveness, highlighting the importance of task graph quality and compatibility for cross-institutional adaptation.

\section{Conclusion}
This work presents \textbf{S}urgical \textbf{P}hase \textbf{A}nywhere (SPA), a lightweight framework that adapts foundation models for cross-institutional surgical workflow understanding with minimal annotation. By combining few-shot spatial adaptation, temporal adaptation, and dynamic test-time adaptation, SPA effectively handles domain shifts and ensures reliable phase recognition. SPA enables scalable, rapid deployment of customized phase recognition models across diverse institutions. This approach advances data-efficient surgical AI, offering a practical solution for real-world clinical applications.

\begin{credits}
\subsubsection{\ackname} This work has received funding from the European Union (ERC, CompSURG, 101088553). Views and opinions expressed are however those of the authors only and do not necessarily reflect those of the European Union or the European Research Council. Neither the European Union nor the granting authority can be held responsible for them. This work was also partially supported by French state funds managed by the ANR under Grants ANR-10-IAHU-02. The authors would like to acknowledge the High Performance Computing Center of the University of Strasbourg for supporting this work by providing scientific support and access to computing resources. Part of the computing resources were funded by the Equipex Equip@Meso project (Programme Investissements d'Avenir) and the CPER Alsacalcul/Big Data.

\subsubsection{\discintname}
The authors have no competing interests to declare that are relevant to the content of this article.
\end{credits}

\bibliographystyle{splncs04}
\bibliography{mybibliography}

\end{document}